\documentclass[runningheads]{llncs}

\usepackage[T1]{fontenc}
\usepackage{graphicx}
\usepackage{hyperref}
\usepackage{color}

\usepackage{amsmath,amssymb,amsfonts}
\usepackage[ruled,noend,linesnumbered]{algorithm2e}
\usepackage{tabularx}
\usepackage{booktabs}
\usepackage{cleveref}
\usepackage{color}
\usepackage{caption}
\usepackage{subfigure}
\usepackage{txfonts}
\usepackage{float}

\begin{document}

\title{Im2win: An Efficient Convolution Paradigm on GPU}

\author{
Shuai Lu\inst{1} \and
Jun Chu\inst{1} \and
Luanzheng Guo\inst{2} \and
Xu T. Liu\inst{3}\orcidID{0000-0003-3980-9803}
}

\institute{Nanchang Hangkong University, Nanchang, Jiangxi Province, China \\
\email{2016085400101@stu.nchu.edu.cn, chuj@nchu.edu.cn} \and
University of California Merced, California, USA \\
\email{lguo4@ucmerced.edu} \and
University of Washington, Seattle, Washington, USA \\
\email{x0@uw.edu}
}

\maketitle

\begin{abstract}
Convolution is the most time-consuming operation in deep neural network operations, so its performance is critical to the overall performance of the neural network. The commonly used methods for convolution on GPU include the general matrix multiplication (GEMM)-based convolution and the direct convolution. GEMM-based convolution relies on the im2col algorithm, which results in a large memory footprint and reduced performance. Direct convolution does not have the large memory footprint problem, but the performance is not on par with GEMM-based approach because of the discontinuous memory access. This paper proposes a window-order-based convolution paradigm on GPU, called \texttt{im2win}, which not only reduces memory footprint but also offers continuous memory accesses, resulting in improved performance. Furthermore, we apply a range of optimization techniques on the convolution CUDA kernel, including shared memory, tiling, micro-kernel, double buffer, and prefetching. We compare our implementation with the direct convolution, and PyTorch's GEMM-based convolution with cuBLAS and six cuDNN-based convolution implementations, with twelve state-of-the-art DNN benchmarks. The experimental results show that our implementation 1) uses less memory footprint by 23.1\% and achieves 3.5$\times$ TFLOPS compared with cuBLAS, 2) uses less memory footprint by 32.8\% and achieves up to 1.8$\times$ TFLOPS compared with the best performant convolutions in cuDNN, and 3) achieves up to 155$\times$ TFLOPS compared with the direct convolution. We further perform an ablation study on the applied optimization techniques and find that the micro-kernel has the greatest positive impact on performance.
\end{abstract}

\section{Introduction}
Convolutional neural network (CNN) is an important network model widely used in computer vision, image processing, and scientific computing. CNN  consists of an input layer, an output layer, and convolutional layers between them~\cite{crowley2018moonshine}. Convolutional operations can take 50\%~-~90\% of the total inference operations of the neural network model~\cite{shufflenet}. Also, convolution operations often account for over 90\% of the total execution time of many neural networks~\cite{efficient_processing_of_DNN}. Therefore, it is critical to reduce the cost of convolutional operations to improve the overall performance of neural networks.

Graphics processing unit (GPU) has been used to accelerate tensor convolution operations. Popular deep learning frameworks, such as PyTorch~\cite{pytorch_nips_2019} and TensorFlow~\cite{tensorflow2015-whitepaper}, use GPU to accelerate convolution operations with cuBLAS~\cite{nvidia2008cublas} and cuDNN~\cite{cudnn_arxiv_2014}, both developed by NVIDIA. cuBLAS is a GPU-accelerated library for the basic linear algebra subroutines. cuDNN is a set of primitives for forward and backward convolution, pooling, normalization, and activation layers used by neural networks.

There are mainly two types of convolution methods on GPU in terms of data transformation: the im2col data transformation-based and no data transformation at all. The im2col-based convolution transforms the input tensor and the filter tensor into two matrices, known as \textit{the im2col algorithm}, followed by the general matrix-matrix multiplication (GEMM) with cuBLAS or cuDNN, and finally transforms the resultant matrix back to the output tensor~\cite{chellapilla_high_2006}. The problem with the im2col-based convolution is that 1) the im2col operation generates a high memory footprint and bandwidth overhead, which is exaggerated on GPU where the memory/cache capacity is highly limited; 2) its performance is significantly affected by the performance of the GEMM operation in cuBLAS, which takes the input im2col matrix and the filter im2col matrix as inputs while the two matrices are significantly different in size, leading to bad performance~\cite{directconvolutions,a_family_of_MM}. 

A typical direct convolution has no data transformation, and is implemented as seven nested for loops over the original input and filter tensors, with the scalar ${a}$ multiplied by ${x}$ plus ${y}$ (AXPY) computed in the innermost loop~\cite{directconvolutions}. Compared to the im2col-based convolution, the direct convolution has no additional memory overhead. However, its AXPY operations suffer from discontinuous memory access, because of visiting distinct dimensions of the input tensor across the nested for loops. This results in low data reuse and low cache hit rate. This problem is seriously magnified on GPU. 

To solve similar problems on CPU, we previously proposed a novel convolution algorithm, called \texttt{im2win}~\cite{im2win_hpec_2022} (image to window), which rearranges the input tensor in the access order of the dot product windows. In this paper, we evolve the \texttt{im2win} algorithm and develop a memory-efficient and high-performance \texttt{im2win}-based convolution paradigm on GPU. The \texttt{im2win} convolution paradigm first transforms the input tensor into an \texttt{im2win} tensor using the \texttt{im2win} data transformation (see~\Cref{subsection:motivation}). Next, the convolution is implemented as a three-level nested loop structure akin to an implicit GEMM convolution, and the indices of input tensor, filter tensor and output tensor can be mapped tothe three levels of for loops when performing an AXPY operation. Our \texttt{im2win} data transformation algorithm can significantly reduce memory consumption compared to the im2col data transformation. We implement the \texttt{im2win}-based convolution paradigm in CUDA and propose a range of optimization techniques, including tiling, micro-kernel, double buffer, and prefetching.

We compare our implementation with various convolution methods, including the direct convolution, PyTorch's GEMM-based convolution using cuBLAS, and six different cuDNN-based convolution implementations, using twelve different state-of-the-art deep neural network benchmarks. The results of our experiments indicate that our implementation outperforms the others in different aspects. Specifically, it uses less memory by 23.1\% compared to cuBLAS and by 32.8\% compared to the best-performing convolution implementations in cuDNN, while on average achieving 3.5$\times$ and 1.1$\times$ TFLOPS, respectively. Additionally, our implementation achieves up to 155$\times$ TFLOPS compared with the direct convolution. We also conduct an ablation study to understand which optimization technique has the greatest positive impact on performance, and find that the micro-kernel has the most significant effect. We make our code publicly available at \href{https://github.com/seth-lu/Im2win}{https://github.com/seth-lu/Im2win} under \textit{cuda} branch.

To summarize, this paper makes the following \textbf{contributions}:\\
\indent 1) We propose an innovative convolution paradigm on GPU, called \texttt{im2win}-based convolution (\Cref{subsection:im2wingpu}), along with a set of optimizations that are specifically designed to improve its memory efficiency and performance (\Cref{subsection:optimizations}). Our proposed convolution paradigm is shown to be both high-performance and memory-efficient, offering a promising alternative to existing convolution methods on GPU.
\\
\indent 2) We implement our \texttt{im2win}-based convolution in CUDA and compare it with the direct convolution, existing convolution algorithms in cuBLAS and cuDNN. We conduct an experimental evaluation using twelve DNN benchmarks of various dimensions that provides a comprehensive result of our proposed method~(\Cref{subsection:performance}). 
\\
\indent 3) 
We conduct an ablation study on the optimization techniques applied to the proposed im2win-based convolution paradigm, which reveals that the micro-kernel optimization technique has the most significant impact on performance~(see~\Cref{subsection:ablationStudy}).

The rest of paper is organized as follow. \Cref{sec:prelim_and_related} defines the notations used in this paper, reviews existing convolution techniques and related works. \Cref{sec:convolution_paradigm} presents our convolution paradigm on GPU along with a set of optimizations that are specifically designed to improve its memory efficiency and performance. We conduct an experimental evaluation, an ablation study and present the performance and memory usage of different convolution algorithms in~\Cref{sec:experiment}. Finally, we conclude our work in~\Cref{sec:conclusion}.

\section{Preliminaries and Related Work}
\label{sec:prelim_and_related}

In this section, we define the notations used in this paper, review the related works in the direct convolution, the GEMM-based convolution and other convolutions.
\subsection{Notations}

Three main tensor data in the convolution operation are the Input tensor ($\mathcal{I}$), the Filter tensor ($\mathcal{F}$), and the Output tensor ($\mathcal{O}$). These tensors in $NCHW$ layout are expressed as $\mathcal{I}[N_{i}][C_{i}][H_{i}][W_{i}]$, $\mathcal{F}[C_{o}][C_{i}][H_{f}][W_{f}]$ and $\mathcal{O}[N_{i}][C_{o}][H_{o}][W_{o}]$. The convolution is defined as:

\begin{equation}
\begin{array}{r}
\begin{aligned}
\mathcal{O}_{(i, j, m, n)}=\sum_{j=0}^{C_{i}-1} \sum_{m=0}^{H_{f}-1} \sum_{n=0}^{W_{f}-1}\left(\mathcal{I}_{(i, j, m \times s+u, m \times s+v)}\right. 
\left.\times \mathcal{F}_{(j, r, u, v)}\right),
\end{aligned}
\end{array}
\end{equation}
subject to
\begin{equation}
\begin{aligned}
&i~=0,1,..,N_{i}-1, j=0,1,..,C_{o}-1, m=0,1,..,H_{o}-1,\\
&n=0,1,..,W_{o}-1, u=0,1,..,H_{f}-1, v=0,1,..,W_{f}-1,\\
&r~=0,1,..,C_{i}-1. \nonumber
\end{aligned}
\end{equation}

$N_{i}$ is the batch size, $s$ is the stride size, $C_{i}$ and $C_{o}$ are the number of input and output channels, $H_{i/f/o}$ and $W_{i/f/o}$ denote height and width in spatial dimensions.

\subsection{The direct convolution}
The direct convolution is one of the most naive implementations of convolutions. A basic direct convolution has seven nested for loops. The outer four loops iterate over the four dimensions of $\mathcal{O}$, and the inner three loops iterate over $\mathcal{F}$ and $\mathcal{I}$. Each element of $\mathcal{O}$ is computed with an AXPY operation in the innermost loop. These nested loops can be parallelized well on GPU. However, the larger $\mathcal{O}$ is, the less data can fit in the cache. In this case, the direct convolution accesses directly through the global memory. The data access is discontinuous and the latency is high, resulting in poor performance~\cite{directconvolutions}. It has been shown that the performance of the direct convolution can be greatly improved by redesigning specific data layouts and data flows on the GPU~\cite{gpnpu}.

\subsection{The GEMM-based convolution}

\begin{figure}[htb]
\centering
\centerline{\includegraphics[scale=0.5]{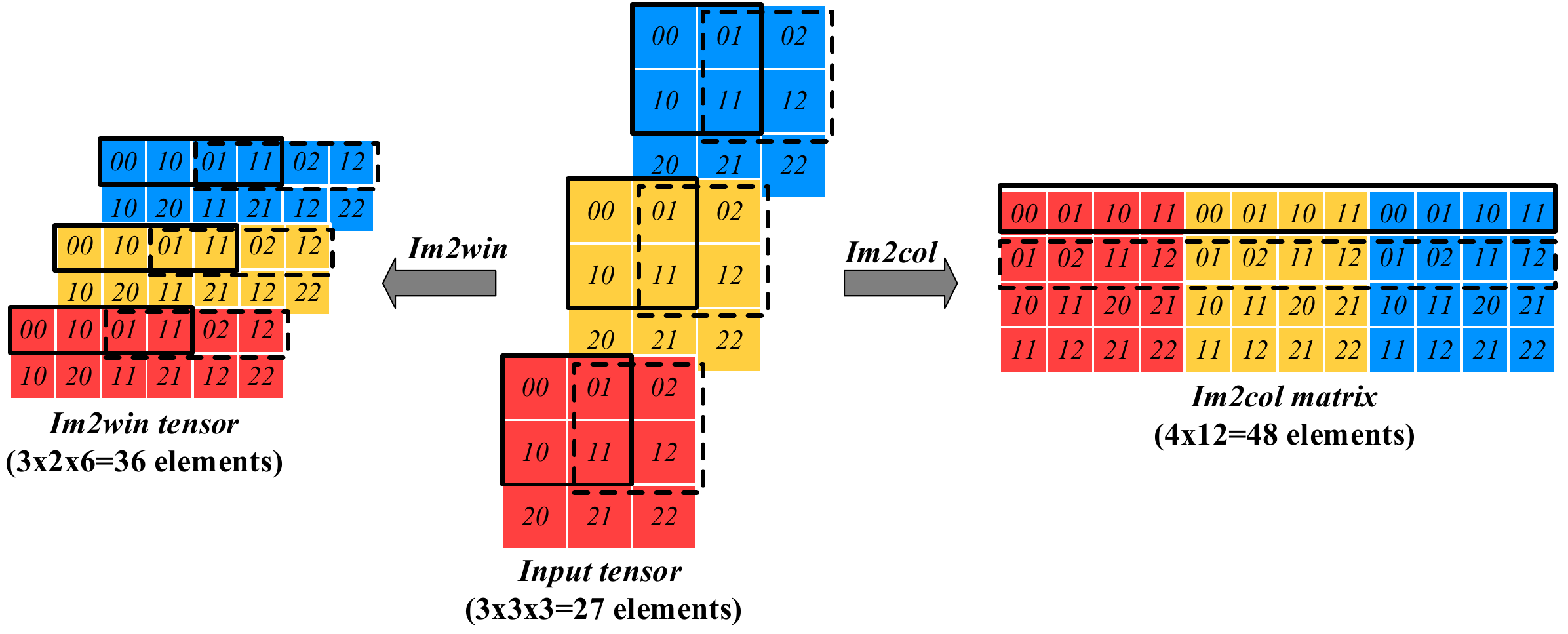}}
\caption{\small The im2col and im2win data transformation examples with $C_{i}$ = $H_{i}$ = $W_{i}$ = 3, $H_{f}$ = $W_{f}$ = 2, $s_{h}$ = $s_{w}$ = 1. The solid and dashed boxes indicate the different dot product windows of the input tensor. We can see that the im2win tensor has less elements than the im2col matrix.}
\label{fig:im2col_and_im2win}
\end{figure}

The GEMM-based convolution proposed by Chellapilla et al.~\cite{chellapilla_high_2006} is the most commonly used convolution algorithm, and is widely used in existing deep learning frameworks~\cite{pytorch_nips_2019,tensorflow2015-whitepaper}. Due to its fundamental and general nature, it is often used as a benchmark for performance comparison. The GEMM-based convolution unrolls the convolution operation into a GEMM operation. The $\mathcal{I}$ of size $N_i \times C_i \times H_i \times W_i$ is processed in $N_i$ batches, each batch contains data $\mathcal{I'}$ of size $C_i \times H_i \times W_i$ (i.e., a single image). As shown on the right in~\Cref{fig:im2col_and_im2win}, the im2col algorithm transforms $\mathcal{I'}$ into a 2D matrix; and $\mathcal{F}$ is unfolded into a filter matrix. In im2col, the elements of each dot product window of $\mathcal{I'}$ is flattened and copied into a single row of a matrix (see~\Cref{fig:im2col_and_im2win}). Denoting the im2col matrix as $M$ and the filter matrix as $N$, the im2col algorithm can be written as: $M(m W_{o}+n, (r H_{f}+u)W_{f}+v)=\mathcal{I'}(r, m+u, n+v),~N((r H_{f}+u)W_{f}+v, j)=\mathcal{F}(j, r, u, v)$. Next, a GEMM operation in BLAS library performs the matrix product of the transformed input matrix and the transformed filter matrix to get the output matrix: $R'=M\times N$. The convolution result tensor $R$ is transformed from $R'$: $R(j,m,n)=R'(m W_{o}+n,j)$.

Dongarra et al. has demonstrated that the GEMM-based convolution benefits from the efficient implementation on GPU and the nature of GPU architectures~\cite{DBLP:conf/iccS/DongarraHHRVZ17}. Due to the highly optimized cuBLAS library, GEMM-based convolution has reliable performance and supports various input tensor sizes. However, this approach requires a large memory to store the im2col matrix transformed from the input tensor and the filter tensor. Because it has to store duplicated elements due to overlap of the filter positions in the convolution, the im2col matrix is much larger than the original tensor. What's worse, the im2col matrix is much larger than the filter matrix, this results the GEMM operation in significantly lower performance than the best achievable performance~\cite{directconvolutions,a_family_of_MM}. MEC proposes a compact lowering trick on the im2col matrix and splits a single GEMM into multiple small GEMMs to reduce the memory footprint~\cite{mec}. The small GEMM operations are performed in parallel to complete the convolution. We intend to compare our convolution with MEC on GPU, unfortunately, MEC is not open-sourced.

\subsection{The convolution algorithms implemented in cuDNN}
cuDNN is a GPU-accelerated deep learning library from NVIDIA, which implements six convolution algorithms including the \textit{direct convolution}, the \textit{GEMM-based convolution}, two \textit{implicit GEMM-based convolutions}, the \textit{Fast Fourier Transform (FFT) convolution}, and the \textit{Winograd convolution}. The implicit GEMM-based convolution is a variant of the direct convolution, which operates natively on the input tensors, converting the computation into a GEMM on the fly. During the computation, the im2col matrices are implicitly formed. There is another variant that precomputes offsets used in the implicit GEMM. The FFT convolution uses the fast Fourier transform to achieve convolution. It can achieve fast convolution with fewer operations the direct convolution, however, it requires more memory and is more difficult to implement as it works with complex numbers instead of real numbers. The Winograd convolution is based on the Winograd's minimal filtering algorithm~\cite{winograd_2016_icpr}, which is computationally efficient for some small convolution kernels.

cuDNN supports autotunning, which automatically selects an algorithm on a per-layer basis based on the layer dimensions. But even so, cuDNN still has some shortcomings. The cuDNN call parameter API is pre-defined, so it does not have the flexibility to build some special convolutions. cuDNN often resorts to a slower algorithm that fits the workspace size constraint. To alleviate this behavior of cuDNN, u-cuDNN divides layers' mini-batch computation into multiple micro-batches transparently by decreasing the workspace size requirements~\cite{ucudnn-cluster-2018}. We refer the readers the performance evaluation of cuDNN convolution algorithms in~\cite{evaluatcudnn_2019}.

\section{The im2win-based convolution paradigm on GPU}
\label{sec:convolution_paradigm}

To reduce the huge memory usage of the im2col-based convolution and avoid nonconsecutive memory access of the direct convolution, we use the \texttt{im2win} data transformation and propose a high-performance and memory efficient \texttt{im2win}-based convolution paradigm on GPU. Furthermore, we propose several optimization techniques for our \texttt{im2win}-based convolution.

\subsection{Motivations} \label{subsection:motivation}

Now we present the \texttt{im2win} data transformation algorithm and the implicit GEMM-based convolution algorithm as the motivations of our \texttt{im2win}-based convolution.

\subsubsection{The \texttt{im2win} data transformation algorithm.}
As shown on the left in~\Cref{fig:im2col_and_im2win}, our image to window algorithm (called \texttt{im2win}) rearranges the input tensor $\mathcal{I}$ in the access order of the dot product windows. It dramatically reduces memory overhead with more compact data arrangement compared with the im2col data transformation algorithm. For the dot product windows in the same row, each dot product operation reuses the elements of the previous loaded window except for the first one. Our im2win algorithm supports great data reusability, temporal and spatial data locality.

In the \texttt{im2win} algorithm, we divide each channel of $\mathcal{I'}$ into $H_o\times W_o$ windows of size $H_f\times W_f$, and copy $W_o$ windows in the same row to one row in our \texttt{im2win} tensor. Performing the above operation for all windows on a single channel of $\mathcal{I'}$, we obtain a tensor of size ($H_o $, $H_f \times W_i$) (see~\Cref{fig:im2col_and_im2win}). This tensor is ordered by the dot product windows and has fewer redundant elements than what the im2col matrix has. Performing the above algorithm for the batch and channel dimensions in $\mathcal{I}$, we will get a tensor of size ($N_i$, $C_i$, $H_o $, $W_i \times H_f$) and call this tensor as an \texttt{im2win} tensor.
Denoting the \texttt{im2win} tensor as $\mathcal{\hat I}$, the algorithm can be written as: 
\begin{equation}
\label{equation:im2win}
\begin{aligned}
\mathcal{\hat I}(i, r, m, k H_{f}+u)=\mathcal{I}(i, r, m+u, n+v).
\end{aligned}
\end{equation}
subject to
\begin{equation}
\begin{aligned}
&m=0,1,..,H_{o}-1,  n=0,1,..,W_{o}-1, u=0,1,..,H_{f}-1,\\
&v~=0,1,..,W_{f}-1, i=0,1,..,N_{i}-1, r=0,1,..,C_{i}-1,\\
&k~=0,1,..,W_{i}-1. \nonumber
\end{aligned}
\end{equation}
Recall in~\Cref{fig:im2col_and_im2win} $s=1$, the im2col matrix has 48 elements, while in~\Cref{fig:im2col_and_im2win}, the im2win tensor has 36 elements. The im2win tensor has $1/3$ less elements than the im2col matrix in addition to provide better data locality and data reusability.

\subsubsection{The implicit GEMM-based convolution algorithm.}
In addition to the GEMM-based convolution algorithm with explicit im2col data transformation, there is also an implicit GEMM-based convolution algorithm, shown in~\Cref{algorithm:IMplicit_GEMM}. Instead of an explicit data transformation process, a three-level nested for loop structure is used in the algorithm to calculate the indices of $\mathcal{I}$~(\Cref{algorithmGemm:line:o_n}~and~\Cref{algorithmGemm:line:f_c}~-~\Cref{algorithmGemm:line:i_w}), $\mathcal{F}$~(\Cref{algorithmGemm:line:o_c}~and~\Cref{algorithmGemm:line:f_c}~-~\Cref{algorithmGemm:line:f_w}) and $\mathcal{O}$~(\Cref{algorithmGemm:line:o_c}~-~\Cref{algorithmGemm:line:o_w}). In the innermost loop, the AXPY operation is performed to result in $\mathcal{O}$~(\Cref{algorithmGemm:line:axpy}). Implicit GEMM-based convolution does not have the memory consumption of data transformation. The name of implicit GEMM-based convolution algorithm can be confusing. With no explicit input and filter matrices, it is impossible to call cuBLAS GEMM API. In addition, the indices to perform an AXPY must be computed on the fly. This algorithm is commonly viewed as a variant of the direct convolution. Since \Cref{algorithm:IMplicit_GEMM} has the same three-level nested for loop structure as GEMM operation, the optimization techniques that are proposed for GEMM can also be applied to implicit GEMM-based convolution algorithm, such as shared memory, tiling, micro-kernel, vectorized load/store and prefetching.

\begin{figure}[htbp]
\begin{minipage}[t]{0.48\textwidth}
\centering
{
\small
\begin{algorithm}[H]
\SetAlgoNoLine
    \caption{Implicit GEMM-based Convolution Algorithm}
    \label{algorithm:IMplicit_GEMM}
    \KwIn{Input $\mathcal{I}$, Filter $\mathcal{F}$, Stride $s$}
    \KwOut{Output $\mathcal{O}$}
    
    \SetKwInOut{dimensions}{Dimensions}
    \dimensions{$\textbf{M} = C_o, \textbf{N} = N_o \times H_o \times W_o, \textbf{K} = C_f \times H_f \times W_f$}
    
    \For{$m = 0$ \bf{to} $M - 1$}{
        $o_c = f_n = m$\\ \label{algorithmGemm:line:o_c}
        \For{$n = 0$ \bf{to} $N - 1$}{    
            $o_n = i_n = n / (H_o \times W_o)$\\ \label{algorithmGemm:line:o_n}
            $o_h = (n \% (H_o \times W_o)) / W_o$\\ \label{algorithmGemm:line:o_h}
            $o_w = (n \% (H_o \times W_o)) \% W_o$\\ \label{algorithmGemm:line:o_w}
            \For{$k = 0$ \bf{to} $K - 1$}{ 
                $f_c = i_c = k / (H_f \times W_f)$\\ \label{algorithmGemm:line:f_c}
                $k_{res} = k \% (H_f \times W_f)$\\
                $f_h = k_{res} / W_f$\\ \label{algorithmGemm:line:f_h}
                $f_w = k_{res} \% W_f$\\ \label{algorithmGemm:line:f_w}
                $i_h = o_h \times s + f_h$\\ \label{algorithmGemm:line:i_h}
                $i_w = o_w \times s + f_w$\\ \label{algorithmGemm:line:i_w}
                $\mathcal{O}(o_n, o_c, o_h, o_w) += \mathcal{I}(i_n, i_c, i_h, i_w) \times \mathcal{F}(f_n, f_c, f_h, f_w)$\\ \label{algorithmGemm:line:axpy}
            }
        }
    }
\end{algorithm}
}
\end{minipage}
\hfill
\begin{minipage}[t]{0.48\textwidth}
\centering
{
\small
\begin{algorithm}[H]
\SetAlgoNoLine
    \caption{Basic Im2win-based Convolution On GPU}
    \label{algorithm:Basic_in2win_convolution_GPU}
    \KwIn{Input $\mathcal{I}$, Filter $\mathcal{F}$, Stride $s$}
    \KwOut{Output $\mathcal{O}$}
    \SetKwInOut{datatrans}{Im2winTensor}
    \datatrans{$\mathcal{\hat I}$ = \textbf{Function} \textsc{im2win}($\mathcal{I},\mathcal{F},s$)}
    
    \SetKwInOut{dimensions}{Dimensions}
    \dimensions{$\textbf{M} = C_o, \textbf{N} = N_o \times H_o \times W_o, \textbf{K} = C_f \times H_f \times W_f$}
    
    \SetKwInOut{blocks}{\# of blocks}
    \blocks{$M/32 \times N/32$}
    
    \SetKwInOut{threads}{\# of threads per block}
    \threads{$32 \times 32$}
    
    $m = bx \times 32 + tx$, 
    $n = by \times 32 + ty$\\ \label{algoBasic:line:compure_m}
    
    $o_c = m$,
    $o_n = i_n = n / (H_o \times W_o)$\\ \label{algoBasic:line:o_c}
    $o_h = (n \% (H_o \times W_o)) / W_o$\\
    $o_w = (n \% (H_o \times W_o)) \% W_o$\\ \label{algoBasic:line:o_w}
    \For{$k = 0$ \bf{to} $K - 1$}{ \label{algoBasic:line:loop_k}
        $f_c = i_c = k / (H_f \times W_f)$\\ \label{algoBasic:line:f_c}
        $k_{res} = k \% (H_f \times W_f)$\\
        $f_h = k_{res} / W_f$,
        $f_w = k_{res} \% W_f$\\ \label{algoBasic:line:f_w}
        $i_h = o_h \times s + f_h$,
        $i_w = o_w \times s + f_w$\\ \label{algoBasic:line:i_w}
        $\mathcal{O}(o_n, o_c, o_h, o_w) += \mathcal{\hat I}(i_n, i_c, i_h, i_w) \times \mathcal{F}(f_n, f_c, f_h, f_w)$\\ \label{algoBasic:line:axpy}
    }

\end{algorithm}
}
\end{minipage}
\end{figure}

\subsection{The \texttt{im2win}-based convolution on GPU} \label{subsection:im2wingpu}

We propose a basic \texttt{im2win}-based convolution on GPU shown in~\Cref{algorithm:Basic_in2win_convolution_GPU} implemented in CUDA. The input tensor $\mathcal{I}$ is initially transformed into the \texttt{im2win} tensor $\mathcal{\hat I}$ based on~\Cref{equation:im2win}. Next, the convolution is implemented as a three-level nested for loop structure same as the implicit GEMM-based convolution. In~\Cref{algorithm:Basic_in2win_convolution_GPU}, dimension M and dimension N are mapped to grid and block respectively, where each block includes 32x32 threads, i.e., grid = (M/32, N/32), block = (32, 32). The $bx$ and $by$ denote block indices in the x and y dimensions respectively, and $tx$ and $ty$ denote thread indices in the x and y dimensions respectively~(\Cref{algoBasic:line:compure_m}). Within the kernel of each block, the three levels of for loops are $\textbf{M} = C_o, \textbf{N} = N_o \times H_o \times W_o$, and $\textbf{K} = C_f \times H_f \times W_f$. The indices of $\mathcal{\hat I}$~(\Cref{algoBasic:line:o_c}~,~\Cref{algoBasic:line:f_c}~-\Cref{algoBasic:line:i_w}), $\mathcal{F}$~(\Cref{algoBasic:line:f_c}~-\Cref{algoBasic:line:f_w}) and $\mathcal{O}$~(\Cref{algoBasic:line:o_c}~-\Cref{algoBasic:line:o_w}) tensor are computed on the fly within the kernel function. The innermost for loop performs an AXPY operation. 

In the kernel function, we first compute indices m and n from dimension M and dimension N respectively from the global indices of the thread tx and ty~(\Cref{algoBasic:line:compure_m}~in~\Cref{algorithm:Basic_in2win_convolution_GPU}). Next, the indices of the four dimensions of the output tensor required for the AXPY operation are calculated by performing division and remainder operations on m and n~(\Cref{algoBasic:line:o_c}~-~\Cref{algoBasic:line:o_w}). Finally, we compute the remaining indices of $\mathcal{\hat I}$ and $\mathcal{F}$ in a for loop in dimension K, and perform AXPY operations after obtaining all the indices of $\mathcal{O}$, $\mathcal{\hat I}$ and $\mathcal{F}$~(\Cref{algoBasic:line:loop_k}~-~\Cref{algoBasic:line:axpy}).

The most expensive computation in~\Cref{algorithm:Basic_in2win_convolution_GPU} is the AXPY operation at~\Cref{algoBasic:line:axpy}, which requires three read operations and one write operation. On GPU, frequent read and write operations to the global memory have substantial latency. Therefore we need to cache as much data as possible used for AXPY operations into shared memory and registers per block, which have much lower latency. At~\Cref{algoBasic:line:o_c}~-~\Cref{algoBasic:line:o_w} of the algorithm, we divide the index of outputs based on the global id of the thread so that each individual thread is responsible for a separate output. This data partition is obvious, but not computationally efficient. We can use the micro-kernel technique (elaborated shortly) to partition the $M_T \times N_T$ of $\mathcal{O}$ computation tasks for each individual thread, which will increase the data reusability. We propose in the next subsection a composition of optimizations making the best use of the \texttt{im2win}-based convolution on GPU. 

{
\small
\begin{algorithm}[t]
\SetAlgoNoLine
    \caption{High Performance Im2win Convolution Algorithm On GPU
    }
    \label{algorithm:High_performance_in2win_convolution_GPU}
    \KwIn{Input tensor $\mathcal{I}$, Filter tensor $\mathcal{F}$, Stride $s$}
    \KwOut{Output tensor $\mathcal{O}$}
    \SetKwInOut{datatrans}{Im2winTensor}
    \datatrans{$\mathcal{\hat I}$ = \textbf{Function} \textsc{im2win}($\mathcal{I},\mathcal{F},s$)}
    
    \SetKwInOut{dimensions}{Dimensions}
    \dimensions{$\textbf{M} = C_o, \textbf{N} = N_o \times H_o \times W_o, \textbf{K} = C_f \times H_f \times W_f$}
    
    \SetKwInOut{blocks}{\# of blocks}
    \blocks{$M/M_B \times N/N_B$}
    
    \SetKwInOut{threads}{\# of threads per block}
    \threads{$M_B / M_T \times N_B / N_T$}

    Registers: $R_\mathcal{\hat I}[2][N_T], R_\mathcal{F}[2][M_T], R_\mathcal{O}[M_T \times M_T]$ \textit{//double buffer}\\ \label{algo2:line:init_reg}
    Shared memories: $S_\mathcal{\hat I}[2][K_B \times N_B], S_\mathcal{F}[2][M_B \times K_B]$ \textit{//double buffer}\\ \label{algo2:line:init_smem}
    $S_\mathcal{\hat I}[0][k_B \times n_B]$ $\underleftarrow{\text{load}}$ $k_B \times n_B$ of $\mathcal{\hat I}(0, by)$ \\ \label{algo2:line:load_Si}
    $S_\mathcal{F}[0][m_B \times k_B]$ $\underleftarrow{\text{load}}$ $m_B \times k_B$ of $\mathcal{F}(bx, 0)$ \\
    \_\_syncthreads() \\ 
    $R_\mathcal{\hat I}[0][n_T]$ $\underleftarrow{\text{vec\_load}}$ $n_T$ of $S_\mathcal{\hat I}[0][0 \times n_B]$ \\ \label{algo2:line:vec_load_Ri}
    $R_\mathcal{F}[0][m_T]$ $\underleftarrow{\text{vec\_load}}$ $m_T$ of $S_\mathcal{F}[0][m_B \times 0]$ \\ \label{algo2:line:vec_load_Rf}

    \For{$kk=0$ \bf{to} $C_{f} \times H_{f} \times W_{f} / K_{f,b} - 1$}{
        \For{$k'=1$ \bf{to} $K_{f,b} - 1$}{
            $R_\mathcal{\hat I}[load][n_T]$ $\underleftarrow{\text{vec\_load}}$ $n_T$ of $S_\mathcal{\hat I}[store][k' \times n_B]$ \textit{//prefetching}\\ \label{algo2:line:pre_Ri}
            $R_\mathcal{F}[load][m_T]$ $\underleftarrow{\text{vec\_load}}$ $m_T$ of $S_\mathcal{F}[store][m_B \times k']$ \textit{//prefetching}\\ \label{algo2:line:pre_Rf}
            $R_\mathcal{O}[m_T \times n_T]$ += $R_\mathcal{F}[store][m_T] \times R_\mathcal{\hat I}[store][n_T]$ \textit{//micro-kernel}\\ \label{algo2:line:micro1}
        }
        \If{$kk \neq C_{f} \times H_{f} \times W_{f} / K_{f,b} - 1$}{
        $S_\mathcal{\hat I}[load][k_B \times n_B]$ $\underleftarrow{\text{load}}$ $k_B \times n_B$ of $\mathcal{\hat I}(kk+1, by)$ \textit{//prefetching}\\ \label{algo2:line:pre_Si}
        $S_\mathcal{F}[load][m_B \times k_B]$ $\underleftarrow{\text{load}}$ $m_B \times k_B$ of $\mathcal{F}(bx, kk+1)$ \textit{//prefetching}\\ \label{algo2:line:pre_Sf}
        \_\_syncthreads() \\
        }
        $R_\mathcal{\hat I}[0][n_T]$ $\underleftarrow{\text{vec\_load}}$ $n_T$ of $S_\mathcal{\hat I}[store][0 \times n_B]$ \\
        $R_\mathcal{F}[0][m_T]$ $\underleftarrow{\text{vec\_load}}$ $m_T$ of $S_\mathcal{F}[store][m_B \times 0]$ \\
        $R_\mathcal{O}[m_T \times n_T]$ += $R_\mathcal{\hat I}[1][n_T] \times R_\mathcal{\hat I}[1][n_T]$ \textit{//micro-kernel}\\ \label{algo2:line:micro2}
    }
    $\mathcal{O}(bx, by)$ $\underleftarrow{\text{store}}$ $R_\mathcal{O}[m_T \times n_T]$

\end{algorithm}
}

\subsection{Optimizations on GPU} \label{subsection:optimizations}
Inspired by the optimization techniques used in GEMM on GPU, we apply the following optimizations to~\Cref{algorithm:Basic_in2win_convolution_GPU}, including tiling, shared memory, micro-kernel, vectorized load/store, double buffer, and prefetching. Those optimizations are especially important to maximize workload and data parallelism and reduce data access latency. We present our high-performance \texttt{im2win}-based convolution on GPU as~\Cref{algorithm:High_performance_in2win_convolution_GPU}.

\textbf{Tiling.} 
Since~\Cref{algorithm:Basic_in2win_convolution_GPU} has a similar implicit GEMM-based convolution implementation with three nested loops of M, N and K, the indices of the input tensor can be divided into small blocks called \textit{tiles}~\cite{tiling}. We tile the sizes of $\mathcal{\hat I}$ and $\mathcal{F}$ into sizes of $M_B \times N_B \times K_B$ at the block level and $M_T \times N_T$ at the thread level in~\Cref{algorithm:High_performance_in2win_convolution_GPU}. As the basic computational unit during computation, the main effect of tiling is to improve computational performance by reducing data accesses and improving data locality. For example, the size of the tile can be adapted to match the size of the shared memory or the registers, which has substantial lower latency, to improve the data reuse and to increase cache hit rate.

\textbf{Shared memory and register.}
The memory on a GPU device consists of four levels of hierarchy: the global memory, the shared memory, the L1\&L2 caches (not programmable in CUDA) and the registers. From the global memory to the shared memory, and to the registers, the access latency decreases and the size also decreases. After tiling the input tensor and the filter tensor, we allocate registers and shared memory blocks of size $M_B \times K_B$ and size $K_B \times N_B$~(\Cref{algo2:line:init_reg}~-~\Cref{algo2:line:init_smem}), and we load $\mathcal{\hat I}$ and $\mathcal{F}$ located in global memory into the registers and shared memory~(\Cref{algo2:line:load_Si}~-~\Cref{algo2:line:vec_load_Rf}) in~\Cref{algorithm:High_performance_in2win_convolution_GPU}. Because each dot product operation reuses the elements of the previous loaded dot product window from the same row in the \texttt{im2win} tensor. To take advantage of this, we load the data to the share memory of each block with as many dot product windows from the same row as possible, to achieve highest possible data reusability and data locality.

\textbf{Micro-kernel.}
The micro-kernel technique can be used to increase the computational intensity. Without it, one AXPY operation in the innermost for loop of our \texttt{im2win}-based convolution computes one element of $\mathcal{O}$. Micro-kernel are typically implemented as outer product multiplications of vectors. With each micro-kernel used in each thread in a block, each thread is now responsible for computing multiple elements of $\mathcal{O}$. We tile the size of the micro-kernel at $M_T \times N_T$ divided at the thread level~(\Cref{algo2:line:micro1}~-~\Cref{algo2:line:micro2}~in~\Cref{algorithm:High_performance_in2win_convolution_GPU}). The micro-kernel partitions the matrix multiplication among multiple threads, reducing the number of memory accesses and improving the parallelism and computational efficiency of the AXPY operations.

\textbf{Vectorized load/store.}
The vectorized load/store are techniques to improve memory access efficiency by loading or storing multiple consecutive data elements from the shared memory into registers under single instruction (SIMD), thereby improving data IO efficiency and memory throughput. Data IO and memory throughput are often the performance bottlenecks when performing convolutional computation on the GPU. Our im2win tensor data structure is stored in a consecutive physical memory, with the dot product windows of the same row arranged continuously. Because each APXY operation loads consecutive dot product windows in the micro-kernel, loading $\mathcal{\hat I}$ and $\mathcal{F}$ from the shared memory of a block into the registers can be done using vectorized load~(\Cref{algo2:line:vec_load_Ri}~-~\Cref{algo2:line:vec_load_Rf}~in~\Cref{algorithm:High_performance_in2win_convolution_GPU}).

\textbf{Double buffer and prefetching.}
The double buffer optimization refers to the use of two buffers to store the input and filter tensors for pipelined concurrent computation. In~\Cref{algorithm:High_performance_in2win_convolution_GPU}, we allocate two registers at~\Cref{algo2:line:init_reg} and two shared memories at~\Cref{algo2:line:init_smem}. Typically, one buffer is used for the ongoing computation and the other is for prefetching the new data used into registers (or shared memory) in the next computation. It hides the latency and overhead of loading data. When the computation is completed, the roles of the two buffers are swapped, i.e., the original buffer becomes the new load buffer and the original load buffer becomes the new computation buffer. 
The prefetching technique is performed on $\mathcal{\hat I}$ and $\mathcal{F}$~(\Cref{algo2:line:pre_Ri}~-~\Cref{algo2:line:pre_Rf}~and~\Cref{algo2:line:pre_Si}~-~\Cref{algo2:line:pre_Sf}~in~\Cref{algorithm:High_performance_in2win_convolution_GPU}), followed by a \textit{\_\_syncthread()} that synchronizes the data among all the threads of a block performing prefetching. 
The prefetching technique allows certain amount of data (we prefetch 128 elements for the shared memory, and 8 elements for the register in our implementation) to be prefetched before the computation, thus reducing data waiting time and improving computational efficiency~\cite{prefetching}.

\section{Experimental Results}
\label{sec:experiment}

In this section, we compare our \texttt{im2win} convolution algorithm with a naive direct convolution, PyTorch's im2col-based algorithm using cuBLAS and cuDNN's convolution implementations, present the performance results and memory usages of them, and perform an ablation study of our proposed optimization techniques.
\subsection{Experimental Setup}
\noindent\textbf{Platform.} We perform our experiments on a NVIDIA GeForce RTX 3090 GPU which has 24GB memory and is connected to an Intel Xeon Silver 4214 CPU server. \\
\noindent\textbf{Software.} The APIs of cuBLAS and cuDNN are pre-defined and are not available for the \texttt{im2win}-based convolution, so we implement our \texttt{im2win} convolution paradigm using CUDA 11.1. We use the tensor data structure of PyTorch 1.10.0~\cite{pytorchlib} with the single 32bit precision. We list the algorithms we compared, theirs notations, and their descriptions in~\Cref{table:experiment_algos}.

\begin{table}[ht]
\centering
\caption{\small The convolution algorithms used in the experimental evaluations, the notations used in figures, and their implementation details.}
\label{table:experiment_algos}
    \begin{tabular}{l|l}
    \toprule
    \textbf{Notation} & \textbf{Description} \\
    \midrule
    im2col+cuBLAS & the im2col-based convolution in PyTorch using cuBLAS 11.2 \\
    direct        & a naive direct convolution implemented in CUDA 11.1  \\
    cuDNN         & six convolutions in PyTorch using cuDNN 8.0.1 \\
    im2winGPU     & our im2win-based convolution implemented in CUDA 11.1 \\
    \bottomrule
    \end{tabular}
\end{table}

\noindent\textbf{Benchmarks.} 
We aim to check how well our convolution paradigm performs on various convolutional layers in terms of dimensions. However, it is not persuasive if we only benchmark with one neural network model. For example, all the filters in VGG-16~\cite{VGG-16} are $3\times3$, and ResNet-50~\cite{resnet} contains only three different filters in sizes. Hence we select twelve state-of-the-art DNN benchmarks~\cite{mec} in our evalution, including twelve unique convolution layers, Conv1-Conv12 (the parameters are shown in~\Cref{table:benchmarks}).

\begin{table}[ht]
\begin{minipage}[c]{0.48\textwidth}
\centering
\captionsetup{justification=centering,singlelinecheck=false}
\caption{\small Parameters of the twelve DNN benchmarks.}
\label{table:benchmarks}

\resizebox{!}{0.46\linewidth}{
    \begin{tabular}{lrrr}
    \toprule
    \textbf{NAME}&\textbf{INPUT}&\textbf{FILTER, STRIDE}&\textbf{OUTPUT} \\
    & $C_{i} \times H_{i} \times W_{i}$ & $C_{o} \times H_{f} \times W_{f}, s_{h}(s_{w})$&$C_{o} \times H_{o} \times W_{o}$\\
    \midrule
    $\textbf{Conv1}$ & $3\times227\times227$& $96\times11\times11, 4$&$96\times55\times55$\\
    $\textbf{Conv2}$ & $3\times231\times231$& $96\times11\times11, 4$&$96\times56\times56$\\
    $\textbf{Conv3}$ & $3\times227\times227$& $64\times7\times7, 2$&$64\times111\times111$\\
    $\textbf{Conv4}$ & $64\times224\times224$& $64\times7\times7, 2$&$64\times109\times109$\\
    $\textbf{Conv5}$ & $96\times24\times24$& $256\times5\times5, 1$&$256\times20\times20$\\
    $\textbf{Conv6}$ & $256\times12\times12$& $512\times3\times3, 1$&$512\times10\times10$\\
    $\textbf{Conv7}$ & $3\times224\times224$& $64\times3\times3, 1$&$64\times222\times222$\\
    $\textbf{Conv8}$ & $64\times112\times112$& $128\times3\times3, 1$&$128\times110\times110$\\
    $\textbf{Conv9}$ & $64\times56\times56$& $64\times3\times3, 1$&$64\times54\times54$\\
    $\textbf{Conv10}$ & $128\times28\times28$& $128\times3\times3, 1$&$128\times26\times26$\\
    $\textbf{Conv11}$ & $256\times14\times14$& $256\times3\times3, 1$&$256\times12\times12$\\
    $\textbf{Conv12}$ & $512\times7\times7$& $512\times3\times3, 1$&$512\times5\times5$\\
    \bottomrule
    \end{tabular}}
\end{minipage}
\begin{minipage}[c]{0.48\textwidth}
\raggedleft
\captionsetup{justification=raggedleft,singlelinecheck=false}
\caption{\small The fastest algorithms selected by cuDNN automatically on twelve benchmarks.}
\label{table:cuDNN_algorithm}
\resizebox{!}{0.46\linewidth}{
    \begin{tabular}{r}
    \toprule
    \textbf{cuDNN ALGORITHM}\\
    Fastest chosen\\
    \midrule
    $IMPLICIT\_GEMM$\\
    $IMPLICIT\_GEMM$\\
    $IMPLICIT\_GEMM$\\
    $IMPLICIT\_GEMM$\\
    $WINOGRAD$\\
    $IMPLICIT\_GEMM$\\
    $IMPLICIT\_GEMM$\\
    $FFT$\\
    $WINOGRAD$\\
    $WINOGRAD$\\
    $WINOGRAD$\\
    $IMPLICIT\_GEMM$\\
    \bottomrule
    \end{tabular}}

\end{minipage}
\end{table}

\subsection{Performance} \label{subsection:performance}
In the experiments, we use the wall-clock time in
the standard C++ library to measure the runtime of different algorithms. We run each algorithm 100 times and record the best runtime among 100 runs. The batch size of each benchmark input data is 128.

\begin{figure}[htbp]
\centering
\centerline{\includegraphics[scale=0.60]{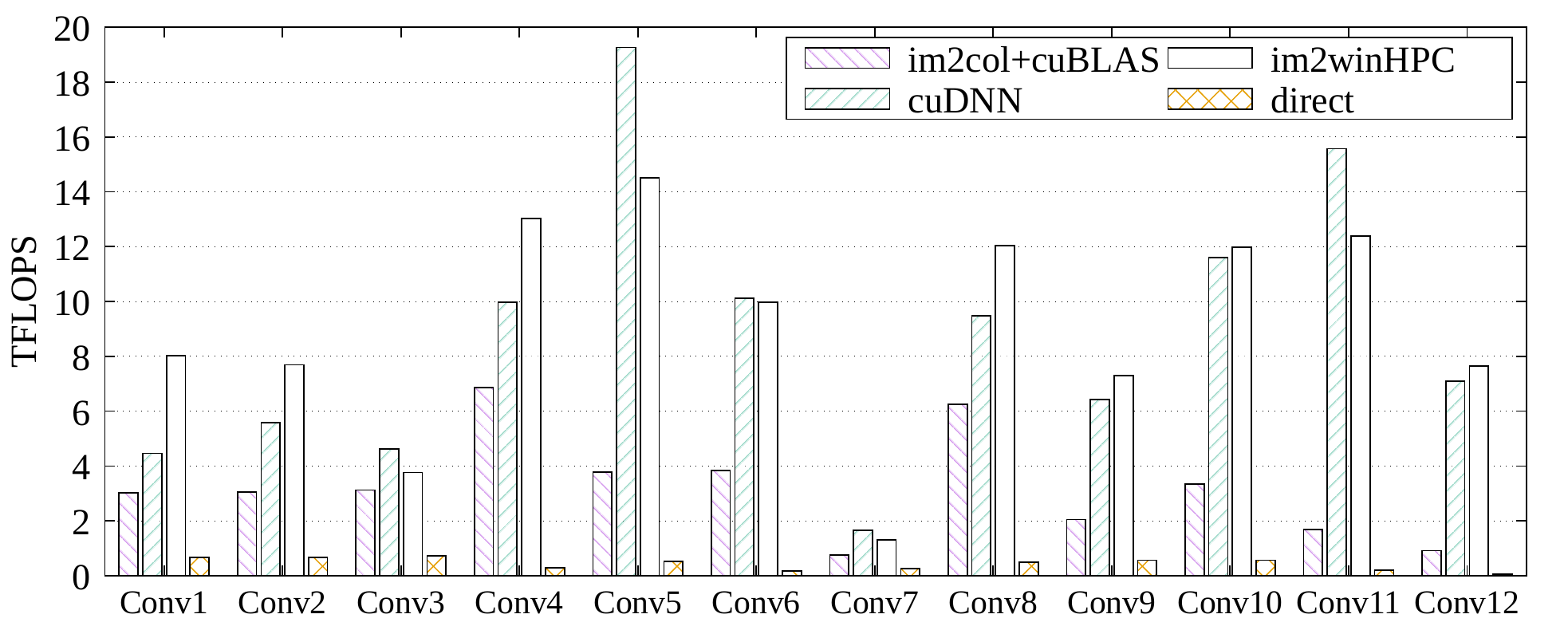}}
\caption{\small Performance comparison of our im2win-based convolution with the direct convolution, the im2col-based convolution using cuBLAS and the convolutions in cuDNN (see~\Cref{table:cuDNN_algorithm}). 
}

\label{fig:TFLOPS}
\end{figure}

\Cref{fig:TFLOPS} shows the TFLOPS of different convolution algorithms of twelve different DNN benchmarks respectively on GPU. cuDNN has six convolution algorithms, with the fastest automatically chosen based on the input tensor dimensions. \Cref{table:cuDNN_algorithm} shows the fastest algorithm automatically chosen by cuDNN at different benchmarks. Among the twelve benchmarks, our im2win-based convolution achieves about on average 3.5 $\times$ TFLOPS than that of im2col+cuBLAS convolution, and achieves 5$\times$ to 155$\times$ TFLOPS compared with the direct convolution Our im2win-based convolution has comparable performance with the cuDNN convolutions and achieves up to 1.8 $\times$ TFLOPS (the first benchmark) than that of the fastest algorithm chosen by cuDNN. Thanks to our customized optimizations tailored for our im2win-based convolution on GPU, we demonstrate better performance than the im2col-based convolution and the direct convolution of cuDNN, and show comparable performance with the implicit GEMM-based convolution, the FFT convolution, and the Winograd convolution in cuDNN.

\begin{figure}[htbp]
\centering
\centerline{\includegraphics[scale=0.60]{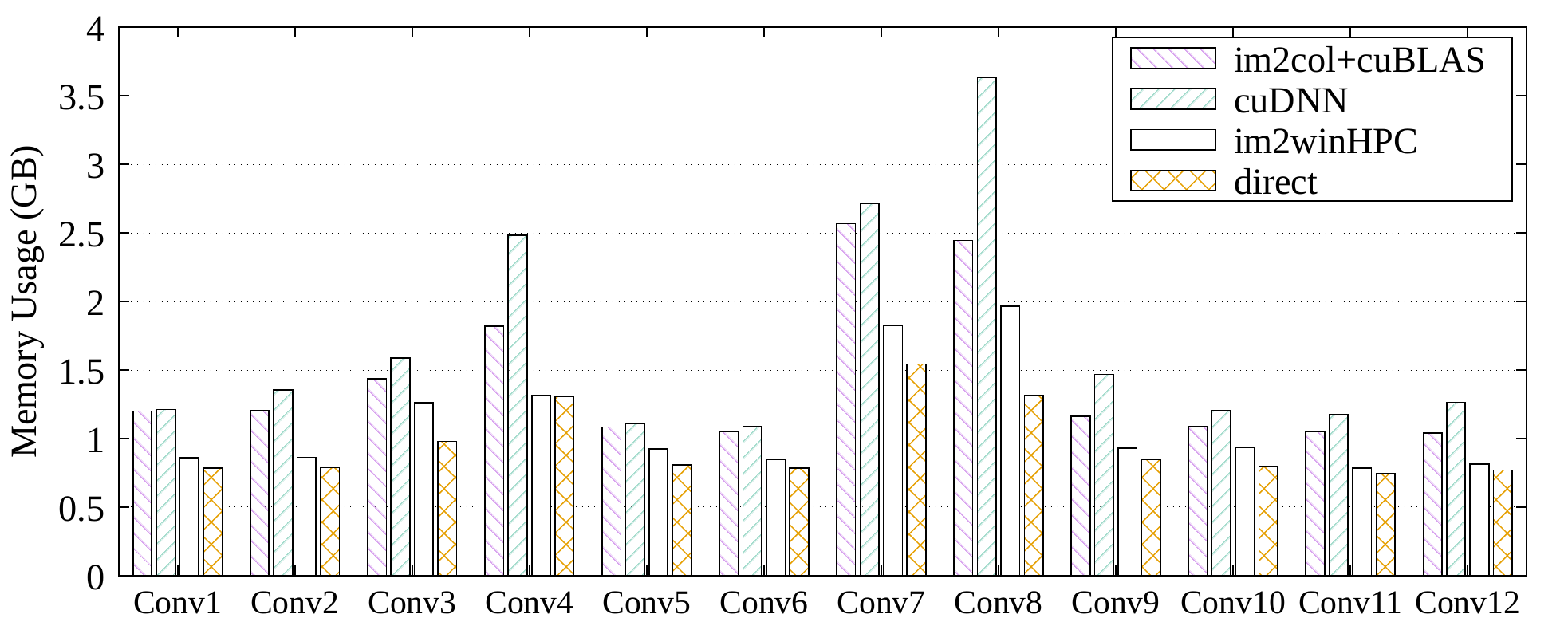}}
\caption{\small Memory usages of our convolution compared to the direct convolution as well as PyTorch's im2col+cuBLAS convolution and cuDNN convolutions. }

\label{fig:memory}
\end{figure}

\begin{figure}[htbp]
\centering
\centerline{\includegraphics[scale=0.60]{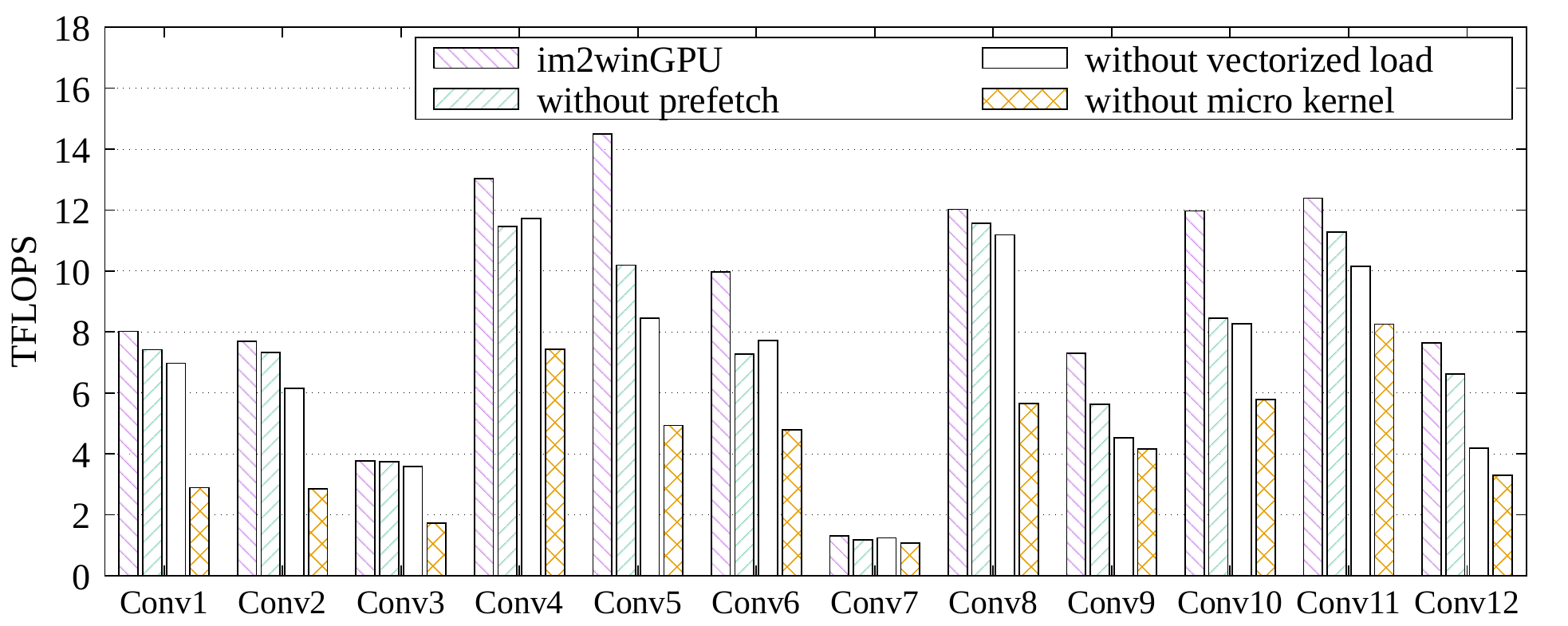}}
\caption{\small Performance comparison of the ablation study on the prefetching, the vectorized load, and the micro-kernel optimization techniques. One technique is removed at a time.}

\label{fig:im2win_Ablation}
\end{figure}

\subsection{Memory Usage} \label{subsection:mem}
\Cref{fig:memory} shows the memory usages of different convolution algorithms on twelve different DNN benchmarks respectively on GPU. Note that cuDNN auto-tunes itself to use the fastest algorithms among its six convolution algorithms based on the input tensor dimensions. The figure shows that our im2win-based convolution algorithm dominantly uses less memory footprint over all twelve benchmarks compared with the im2col-based convolution in cuBLAS and the fastest convolution among the six algorithms in cuDNN. On average, our algorithm uses 23.1\% less memory than cuBLAS, and uses 32.8\% less memory than cuDNN. Our algorithm has slightly higher memory usage than the direct convolution. Considering that the memory of a single GPU is usually not big (even Nvidia A100 has at most 80GB of memory), our convolution paradigm supports substantially larger tensor to be processed on a single GPU over cuBLAS and cuDNN, which is much preferable. 

\subsection{Ablation Study} \label{subsection:ablationStudy}
To explore the performance impact of the prefetching (along with double buffer), the vectorized load, and the micro-kernel techniques we apply in our kernel, we conduct an ablation study on our high-performance \texttt{im2win}-based convolution paradigm. We have im2winGPU as the baseline, which includes all the optimization techniques. For other three variants, we remove one technique at a time to study its effectiveness.
\Cref{fig:im2win_Ablation} shows the performance impact of different optimization techniques on our convolution paradigm in terms of the TFLOPS metric. 
Among the twelve benchmarks, the micro-kernel technique gives the greatest performance boost, followed by the vectorized load, and the prefetching gives the poorest performance boost for our paradigm.

With the micro-kernel implemented as outer product multiplications of vectors in a thread of a block, each thread computes multiple elements of the output tensor $\mathcal{O}$ instead of one. This reduces the number of memory accesses and improves the parallelism and computational intensity of the AXPY operations.
The vectorized load improve data IO efficiency and memory throughput by loading or storing multiple contiguous data elements from the shared memory into the register.
Allocating two buffers (one for prefetching, the other for computation) cuts the size of the available shared memory and registers during computation by half, resulting minimal performance improvement.

\section{Conclusion}
\label{sec:conclusion}

In this paper, we proposed a new convolution paradigm on GPU. We implemented a window-order-based convolution (called im2win) on GPU using CUDA along with a range of optimizations, including shared memory, tiling, micro-kernel, double buffer, and prefetching. 
Using twelve DNN benchmarks, we compared our algorithm with the direct convolution, PyTorch's GEMM-based convolution implementation in cuBLAS and six convolution algorithms in cuDNN. 
The experimental results demonstrate the superior memory and performance efficiency of our im2win-based convolution paradigm compared with the direct convolution and the im2col-based convolution and show comparable performance with the implicit GEMM-based convolution, the FFT convolution, and the Winograd convolution in cuDNN with much less memory footprint. 

\subsubsection{Acknowledgment} 
This work was partly supported by National Natural Science Foundation of China (Grant No. 62162045), Key Research and Development Program of Jiangxi (Program No. 20192BBE50073), and Technology Innovation Guidance Program Project of Jiangxi Province (Special Project of Technology Cooperation) (Grant No. 20212BDH81003).

\bibliographystyle{splncs04}
\bibliography{bibliography.bib}

\end{document}